\documentclass{article}
\usepackage{spconf}
\usepackage{adjustbox, amsmath, amssymb, graphicx, multirow, booktabs}
\usepackage{algorithm}
\usepackage{algpseudocode}
\usepackage{url}
\usepackage{subfigure}
\usepackage{IEEEtrantools}
\makeatletter
\providecommand\add@text{}
\newcommand\tagaddtext[1]{%
  \gdef\add@text{#1\gdef\add@text{}}}%

\renewcommand\tagform@[1]{%
  \maketag@@@{\llap{\add@text\quad}(\ignorespaces#1\unskip\@@italiccorr)}%
}
\makeatother

\usepackage{pgfplots}
\pgfplotsset{grid style={dashed,gray}}
\pgfplotsset{compat=1.12}
\usepackage{tikz}
\usetikzlibrary{shapes.geometric}
\usetikzlibrary{shapes.misc,shadows}
\usetikzlibrary{positioning} % ...positioning nodes
\usetikzlibrary{arrows} % ...customizing arrows
\usetikzlibrary{arrows.meta}
\tikzset{%
    >={Latex[width=1mm,length=1mm]},
    base/.style = {
        rectangle, rounded corners, draw=black,
        minimum width=2cm, minimum height=.4cm,
        text centered, font=\tiny},
    acoustic_model/.style = {base, fill=red!15},
    language_model/.style = {base, fill=cyan!20},
    joint/.style = {base, fill=yellow!15},
    io/.style = {base, fill=none, draw=none, minimum width=0cm},
    data/.style = {
        rectangle, draw, align=center, left color=blue!20, right color=white,
        minimum width=0.5cm, minimum height=0.5cm},
    block/.style ={
        rectangle, thick, draw=black, align=center, fill=orange!15,
        minimum height=.3cm, minimum width=2cm, text width=2cm},
    connector/.style={-latex, font=\tiny},
    rectangle connector/.style={
        connector,
        to path={(\tikztostart) -- ++(#1,0pt) \tikztonodes |- (\tikztotarget) },
        pos=0.5
    },
}

\usepackage{color}

\copyrightnotice{\copyright\ IEEE 2023}

\newif\ifblind
\blindfalse
\title{The Gift of Feedback: Improving ASR Model Quality by Learning from User Corrections through Federated Learning}

\ifblind
    \name{BLIND}
    \address{BLIND}
\else
    \name{\begin{tabular}{c}Lillian Zhou \qquad Yuxin Ding \qquad Mingqing Chen \qquad Harry Zhang \\ \qquad Rohit Prabhavalkar \qquad Dhruv Guliani \qquad Giovanni Motta \qquad Rajiv Mathews \end{tabular}}
    \address{Google LLC, Mountain View, CA, U.S.A. \\
    \{lqz, yxding, mingqing\}@google.com}
\fi

\begin{document}

% \copyrightnotice{979-8-3503-0689-7/23/\$31.00~\copyright2023 IEEE}
\bstctlcite{IEEEexample:BSTcontrol}
\maketitle
\begin{abstract}

Automatic speech recognition (ASR) models are typically trained on large datasets of transcribed speech. As language evolves and new terms come into use, these models can become outdated and stale. In the context of models trained on the server but deployed on edge devices, errors may result from the mismatch between server training data and actual on-device usage. In this work, we seek to continually learn from on-device user corrections through Federated Learning (FL) to address this issue. We explore techniques to target fresh terms that the model has not previously encountered, learn long-tail words, and mitigate catastrophic forgetting. In experimental evaluations, we find that the proposed techniques improve model recognition of fresh terms, while preserving quality on the overall language distribution.

\end{abstract}
\begin{keywords}
speech recognition, federated learning, deep learning, catastrophic forgetting, on-device training
\end{keywords}
\section{Introduction}
\label{sec:intro}

Human language is constantly shifting and evolving. In order to serve a high quality ASR model that can be deployed to user devices as an input mechanism, it is crucial to train on data that is representative of the actual, current vocabulary of user dictation. While it is possible to use proxy data for initial server-side training, stopping there means that the model will lag behind the ever-changing user distribution in terms of freshness and accuracy.

One way to improve model freshness is to make use of the natural feedback loop that occurs during usage. When the model makes recognition errors, users may make manual edits to the output text; in the ideal case, this points to a misrecognition that the model has made, and additionally gives us the corrected transcript to learn from. Paired with the original audio, these training examples are a treasure trove for improving model quality, containing up-to-date transcriptions that faithfully reflect actual user requirements. Federated Learning (FL) \cite{bonawitz2019federated, mcmahan_2017_fl} affords us a privacy-preserving mechanism to leverage this training data and these valuable user correction signals.

Related work has explored mining challenging training examples for model improvement \cite{xue2019hard,Qu_2023,sim2019personalization,alon2018contextual}, but in this work we \textbf{utilize the model's own errors, and user corrections thereof, to target fresh terms}. In this way, the model can improve on precisely the words it struggles with, and learn fresh terms unseen in the training corpus snapshot on the server.

However, not all user edits made naturally in the course of usage are necessarily true corrections to the original spoken utterance. In many cases, the edits may be revisions of original intent, and the resulting edited text may diverge entirely from the original audio. In this work, we \textbf{propose a simple method to target legitimate user corrections: filtering training examples to those that contain terms the model is likely to misrecognize}, for example fresh terms that did not exist when the server training data was collected, and would be unknown to the server model. These misrecognized words are likely to be the target of true user corrections. As shown in our experimental results, this approach successfully helps to target high-quality training examples, and fine-tuning on them improves model quality on these fresh terms.

This improvement on the targeted terms, however, can also come with the unintended pitfall of regression on the original distribution. We \textbf{compare a number of techniques for mitigating catastrophic forgetting~\cite{FRENCH1999128, MCCLOSKEY1989109} in FL}, including variants of weight averaging algorithms~\cite{eeckt2023weight}. We also reintroduce data from the original distribution, which in the FL context requires mixing centralized training in with federated rounds~\cite{augenstein2022mixed}, to positive effect.

Finally, these fresh terms can be long-tail words, and training examples may be exceedingly sparse. We \textbf{propose two approaches to improve learning on the rarest examples}. We find that these combined techniques allow us to learn fresh terms without harming overall model quality.
\section{Methodology}
\label{sec:methodology}

\subsection{Federated Learning}
\label{ssec:fl}

Federated Learning \cite{bonawitz2019federated, mcmahan_2017_fl} is a technique to train a single centralized model in a distributed fashion. During each round of federated training, participating edge devices receive the most recent centralized checkpoint, locally train on available on-device training examples, and send only the model gradients back to the server, never any user data. These updates are aggregated into the central model, which is then sent out again for the next federated round.

In the context of this work, on-device training examples consist of speech audio, the original transcript output by the inference model, along with the final text committed by the user after any edits.

\subsection{Filtering Examples By Fresh Words}
\label{ssec:wordlist}

\begin{figure}[ht]
\centering
\includegraphics[width=.9\columnwidth,bb=0 0 603 295]{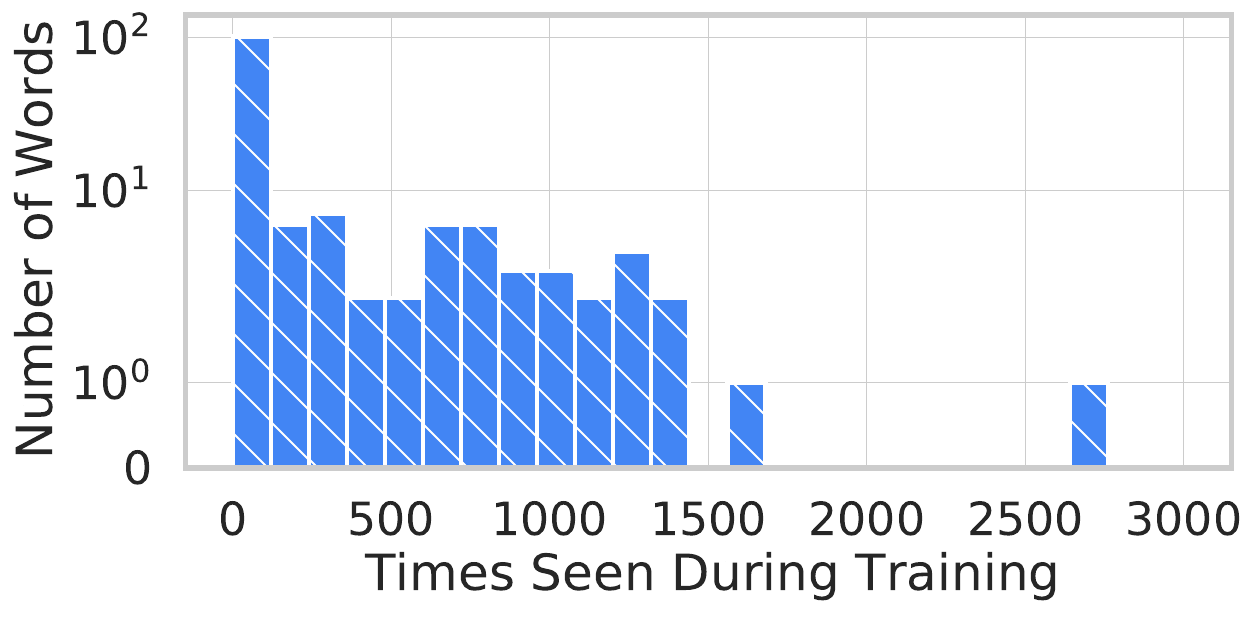}
\caption{Histogram of wordlist word frequency during training. Approximately 50\% of words were seen after 300 rounds of training, with a wide range of frequency.}
\label{fig:freq}
\end{figure}

In order to target high-quality training examples, we generate a list of fresh words that are likely to be misrecognized by the model, and hence the target of legitimate user correction, rather than other edits. In the long term, this step should be automated, e.g., by aggregating over user corrections across devices through Federated Analytics~\cite{zhu2020federated}. Words that are corrected by a large number of users are likely to be true misrecognitions. For this work, we experiment with a manually-curated list of 241 words selected from sources such as pop culture, current events, and recent technologies. The terms vary greatly in how often they appear in on-device training examples (Fig~\ref{fig:freq}), and some examples are shown in Table~\ref{tab:filter_words}.

\begin{table}
    \centering
    \small
    \begin{tabular}{c c c}
        \toprule
        \textbf{Word} & \textbf{Source} & \textbf{Seen Count} \\
        \midrule
        warnock & current events & 85 \\
        webb & technology & 48 \\
        addams & media & 38 \\
        mbappe & sports & 27 \\
        salman & current events & 8 \\
        sumeru & media & 8 \\
        \bottomrule
    \end{tabular}
    \caption{Examples of words used for filtering, based on fresh or trending terms at time of experimentation.}
    \label{tab:filter_words}
\end{table}

While training in FL, we filter the utterances available on-device to examples that contain at least one word from this wordlist and have a user edit. In this way, we target cases where the user spoke one of these words, the model produced the wrong transcript, and the user corrected it.

Because these words are by nature long-tail and represent only a small portion of utterances, evaluating improvement is challenging. To do so, we create a targeted testset focused on measuring quality on these rare words. We compile 4-5 sentences containing each wordlist word: a combination of manually-generated sentences and sentences scraped from the web. Then we generate corresponding audio using Text-to-Speech (TTS)~\cite{oord2017parallel}, and use the resulting audio-sentence pairs as our targeted testset. Due to the nature of TTS audio, and the fact that the terms used for the testset are inherently challenging and rare, including terms that were infrequently or never encountered during training, the resulting WER tends to be high, but serves as a useful benchmark to understand quality improvement on these targeted words. We further perform per-word error analysis before and after fine-tuning to directly understand how recognition of each term is affected.

\subsection{Mitigating Forgetting}
\label{ssec:forgetting}

\begin{figure}[ht]
\centering
    \subfigure[Static checkpoint averaging]{
    \includegraphics[width=.78\columnwidth,bb= 0 0 460 216]{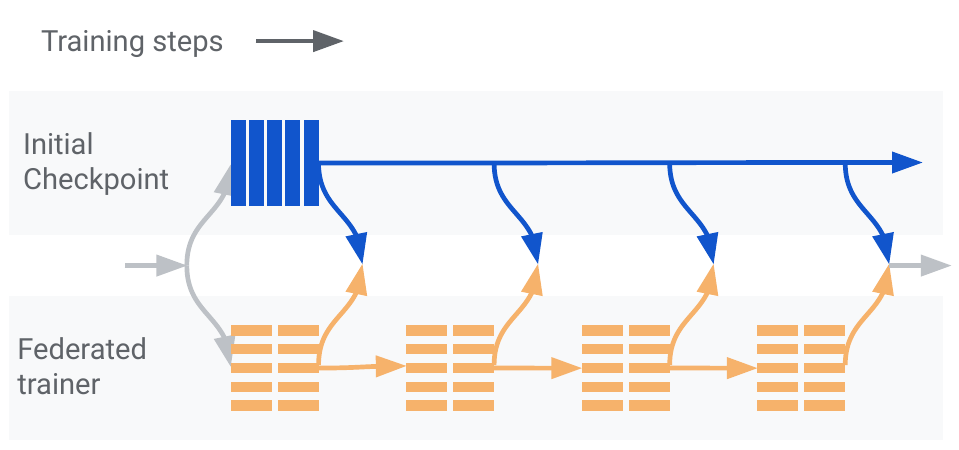}}
    \label{fig:static}
    \subfigure[Dynamic checkpoint averaging]{
    \includegraphics[width=.78\columnwidth,bb= 0 0 460 216]{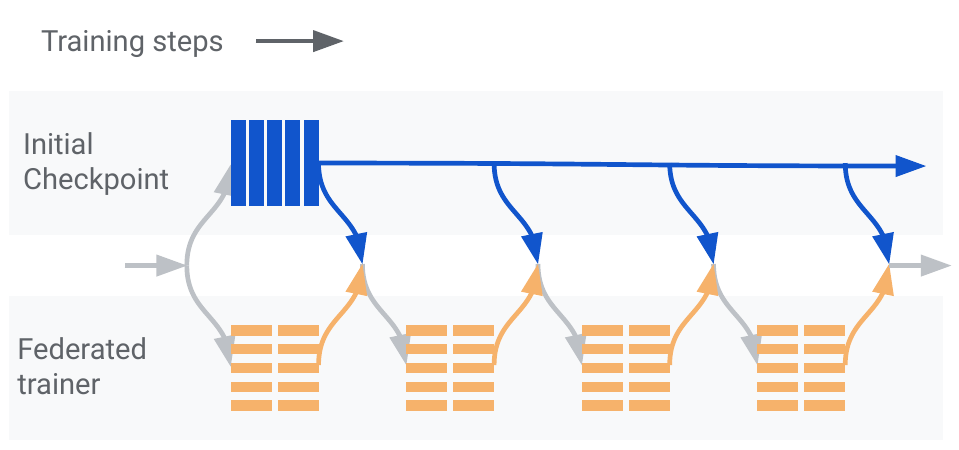}}
    \subfigure[Mixture of Centralized and Federated Training]{
    \includegraphics[width=.78\columnwidth,bb= 0 0 460 216]{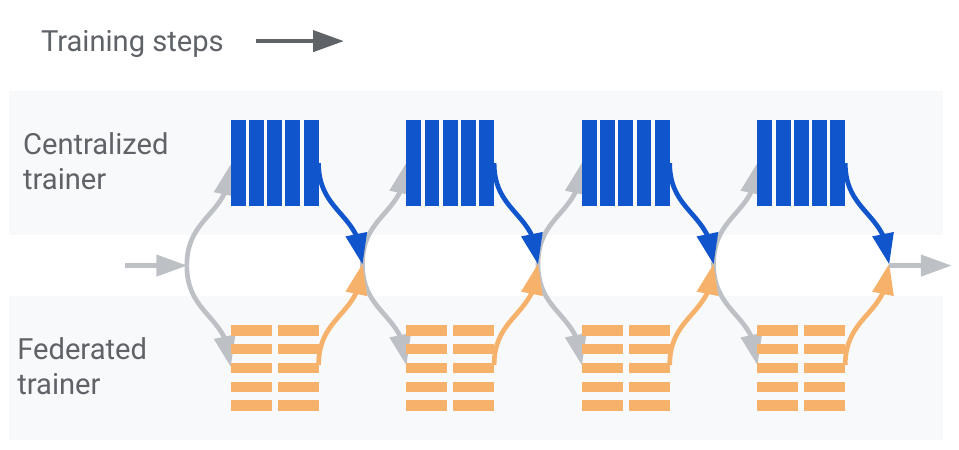}}
\caption{Illustration of techniques to mitigate forgetting. Each shows consecutive federated training rounds (yellow), with varying types of server-side modification (blue), and progression of the centralized model being trained (gray).}
\label{fig:forgetting}
\end{figure}

\subsubsection{Static Checkpoint Averaging}
\label{sssec:staticckptavg}

To mitigate forgetting, we experiment with several techniques, starting with simple weight averaging \cite{eeckt2023weight}. Prior to evaluation, we average the weights of each federated checkpoint with the weights of the initial pre-trained checkpoint, modified by a scaling factor (Fig~\ref{fig:forgetting}a). Federated rounds proceed as usual, and are not impacted by this averaging.

\subsubsection{Dynamic Checkpoint Averaging}
\label{sssec:dynamicckptavg}

As above, this approach averages the weights of each federated checkpoint with the initial checkpoint (Fig~\ref{fig:forgetting}b). However, here each averaged checkpoint is then used to initialize the subsequent federated round. In other words, the federated training process is modified to initialize from the new averaged checkpoint, rather than from the purely federated checkpoint from the round before.

\subsubsection{Mixture of Centralized and Federated Training}
\label{sssec:mixture}

Finally, we perform server-side training in parallel with federated training (Fig~\ref{fig:forgetting}c), as described in \cite{augenstein2022mixed}. While FL rounds learn the on-device data, we simultaneously perform centralized training on the same server datasets used to train the initial pre-trained checkpoint. These centralized model updates are included along with those of the federated clients during aggregation at the end of each round.

\subsection{Learning Long-tail Words}
\label{ssec:longtail}
Because the fresh and misrecognized words tend to be long-tail words, training examples are difficult to come by. All but non-existent in the server training corpus, they are rare even in on-device data. Within the long-tail words, availability also varies drastically from term to term; training on this imbalanced distribution results in little improvement on the least frequent words. To address this, we propose two techniques: probabilistic sampling and client loss weighting.

\subsubsection{Probabilistic Sampling}
\label{sssec:probsamp}

Probabilistic sampling aims to artificially massage the training examples into a more uniform distribution by down-sampling the federated clients that only have examples of the most common words from the wordlist. Each word is assigned a sampling probability $p$, where $p\in(0, 1]$. The more frequent a word is in the original data distribution, the smaller its $p$ value.

If the client data contains a targeted word $w$, then it is included in the FL round with probability $p_{w}$. If the client has multiple wordlist words, either within the same utterance or over multiple training examples, then its probability is the maximum among the words' sampling probabilities. Due to the limited number of clients included in each round, this increases the occurrences of less frequent words seen during training.

\subsubsection{Client Loss Weighting}
\label{sssec:lossweight}

Client loss weighting aims to more heavily penalize the wrong predictions of the rarest wordlist words. Each target word is assigned a client loss weight ($w$). The more frequent the word, the smaller the $w$. On the client, the loss is computed for each utterance, and then additionally scaled by the $w$ of any wordlist word the utterance contains. If an utterance contains multiple target words, its $w$ is the maximum among all the weights. The client loss is the sum of each utterance loss multiplied by the utterance loss weight:

\begin{equation}
\mathcal{L}^{\text{client}} = \sum_{u \in U}\max_{d \in D}(w_{d}) \cdot \text{loss}_{u}
\end{equation}

\noindent
where $\mathcal{L}^{\text{client}}$ is the client loss, computed over $U$ utterances on the client and $D$ words per utterance.
\section{Model and Data}
\label{sec:model}

\subsection{Architecture}
\label{ssec:archictecture}

All models described in this paper are end-to-end, streaming transducer models~\cite{sequence_models_in_asr, e2e_surpasses_server, he2018streaming} based on the Conformer architecture~\cite{gulati_2020_conformer}. Due to the cost of training in FL, we initially demonstrate the wordlist filtering approach in server-side simulation, before expanding to a production FL setup on user devices. For the simulation experiments, we use a streaming Conformer model~\cite{li_2021_better_conformer}, while in production, we use a variation of this model that utilizes cascading encoders~\cite{sainath_2021_nonstreaming_asr}.

One important consideration for on-device learning is computational efficiency. Research in deep learning has shown that neural networks benefit from being overparameterized~\cite{zhu_2018_overparam, neyshabur_2018_overparam}. In particular, when seeking to learn new words in a deep ASR model, previous research has shown the majority of quality gains can be achieved by training only the topmost layers, such as the joint layer, or the joint and prediction network (which together we refer to as the decoder portion of the model)~\cite{sim2019personalization}. We adopt this setting; compared to training the entire model, this affords us significant memory savings. For example, for the production model~\cite{sainath_2021_nonstreaming_asr}, the decoder portion is only 40M parameters out of the entire 150M-parameter model. A static, frozen, and compressed encoder is used to generate encoder features, which are then used as input for decoder training.

\subsubsection{On-Device Minimum Word Error Rate Training}
\label{ssec:mwer}

In addition to standard RNN-t loss, we also incorporate Minimum Word Error Rate (MWER) loss \cite{prabhavalkar2017minimum}, which is computed from the loss of each of the N-best hypotheses from beam search. If $x$ are the input acoustic frames, and $y_n$ are the N-best hypotheses for the output label sequence, then the loss can be written as:

\begin{equation}
\mathcal{L}^{\text{MWER}} = \sum_{i=1}^N \hat{P}(y_i|x) \cdot \left( W(y_i , y) - \frac{\sum_{i=1}^N W(y_i , y)}{N} \right)
\end{equation}

\noindent
where $\hat{P}$ is the probability of the i-th hypothesis normalized over all N-best hypotheses, and $W(y_i , y)$ is the number of word errors in the i-th hypothesis relative to the ground truth.

MWER loss is a component of server-side training, but would be too computationally expensive for an on-device setting due to the memory cost of computing the N-best hypotheses. Instead, we create a modified version for on-device by caching the hypotheses produced by the model decoder during training, and incorporating them during loss computation.

\subsection{Training Data}
\label{ssec:data}

In all our experiments, the model is pretrained on speech data from a multi-domain dataset (MD), encompassing \ifblind (redacted for blind review)\else domains of search, farfield, telephony, YouTube, etc~\cite{misra21_interspeech, recognizing_long_form}\fi, including a Short-Form (SF) and Medium-Form (MF) domain. The model is then fine-tuned on training examples containing the interesting terms. \ifblind\else All data are anonymized, and are collected, managed, and used for training models in accordance with Google AI Principles~\cite{ai_principles}.\fi

For our simulation experiments, the fine-tuning dataset consists of utterances from the SF domain that contain long-tail words (SF-LT). Learning is evaluated on the targeted SF-LT testset, a disjoint set of utterances containing the same long-tail words as the training data. Additionally, to ensure the model quality does not degrade on the overall distribution, we also use disjoint testsets from the SF and MF domains to measure regression on non-targeted words.

In production experiments, a similar pre-trained model is fine-tuned on user data through on-device FL, filtered to examples containing the long-tail words. For server-side evaluation, we create a corresponding targeted eval set, as outlined in the methodology. We also make use of the MF testset as our benchmark for preserving quality on the overall distribution. In both these cases, our goal is to improve WER on the targeted testsets, while maintaining WER on the overall distribution testsets.
\section {Experimental Results}
\label{sec:experiments}

\subsection{In Simulation}
\label{ssec:simulations}

\begin{table}
    \centering
    \small
    \begin{tabular}{c c c c}
        \toprule
        \textbf{Dataset} & \textbf{MF WER} & \textbf{SF WER} & \textbf{SF-LT WER} \\
        \midrule
            MD (Baseline) & 3.7 & 6.3 & 45.5 \\
            \midrule
            \multicolumn{4}{c}{Fine-tuning Whole Model} \\
            \midrule
            SF-LT & \textcolor{red}{7.8} & \textcolor{red}{9.3} & \textbf{31.4} \\
            SF-LT + SF + MF & \textcolor{red}{4.4} & \textbf{6.2} & 35.1 \\
            \midrule
            \multicolumn{4}{c}{Fine-tuning Joint Layer} \\
            \midrule
            SF-LT & \textcolor{red}{6.4} & \textcolor{red}{8.4} & 38.1 \\
            SF-LT + SF + MF & \textbf{3.8} & 6.3 & 40.6 \\
        \bottomrule
    \end{tabular}
    \caption{In simulation, fine-tuning afforded improvement on the targeted testset (SF-LT), but degraded overall testset quality (MF + SF). Reintroducing SF + MF training data mitigated forgetting. When training was limited to the joint layer, the model was still able to improve on the targeted testset, while recovering the original WER for the overall testsets.}
    \label{tab:sim_results}
\end{table}

As shown in Table~\ref{tab:sim_results}, we started with a baseline trained on our multidomain dataset. By fine-tuning only on examples containing the long-tail words (SF-LT), we saw a 31\% relative improvement over the baseline on the targeted testset, but observed catastrophic forgetting in the form of significant degradation on the testsets that represent the overall distribution, including over 100\% regression on MF.

As these simulation experiments were done entirely server-side, we were able to address forgetting by reintroducing training examples from the original domain (SF and MF) during training, in equal proportion to the SF-LT dataset, similar to the Mixture of Centralized and Federated Training approach we proposed for production. This way, we were still able to achieve a 23\% improvement in the targeted testset, while seeing much less degradation in the overall domain.

In order to understand the most memory-efficient setting, we also tried fine-tuning only the joint layer of the model. Here, we saw less improvement on the targeted testset, but even lower degradation on the overall domain. By fine-tuning on both the targeted data and the overall data distributions, we saw a 11\% improvement on the targeted testset, and little or no regression on the overall testsets.

\subsection{Necessity of Wordlist Filtering In Production}
\label{sec:prd}

\begin{figure}[ht]
\centering
\includegraphics[width=.9\columnwidth,bb=0 0 587 374]{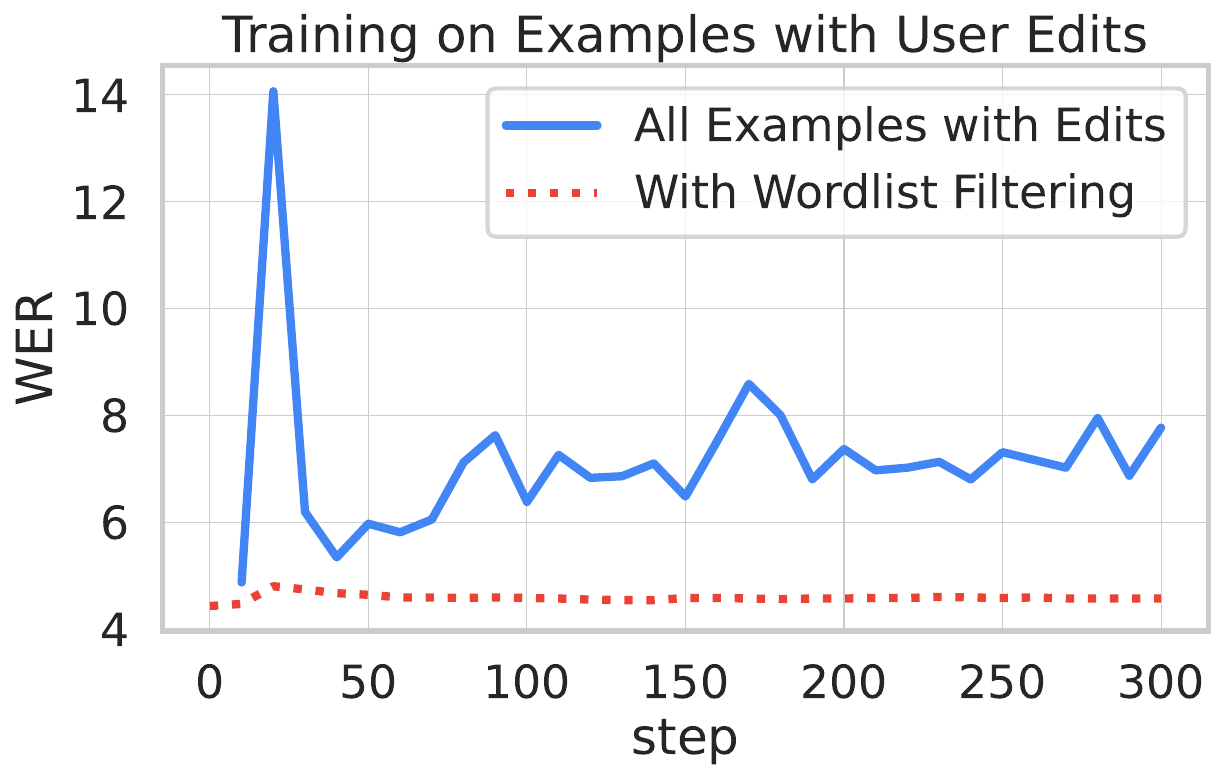}
\caption{Training on all examples with user edits caused the model quality to quickly diverge, indicating many were not actual corrections. However, filtering these examples by our wordlist resulted in far more stable model quality.}
\label{fig:all_user_edits}
\end{figure}

In production experiments, we began by training on all examples with user edits, but the model quality quickly and dramatically degraded (Figure~\ref{fig:all_user_edits}), lending credence to our hypothesis that many of these edits are indeed revisions to the original utterance, rather than corrections of an incorrect ASR transcript, and likely result in significantly divergent text-audio pairs.

However, using our list of fresh words that were likely to be misrecognized, we were able to filter training examples largely to those containing true user corrections. When on-device training was limited to examples with edits that contained a word in our fresh wordlist, the WER was stabilized.

\subsection{Catastrophic Forgetting}
\label{sec:cf}

\begin{table}
    \centering
    \small
    \begin{tabular}{c c c}
        \toprule
        \textbf{Setup} & \textbf{Overall WER} & \textbf{Targeted WER} \\
        \midrule
            Baseline & 4.4 & 17.5 \\
            Pure FL & \textcolor{red}{4.6} & \textbf{16.1} \\
            Static Ckpt Avg & \textbf{4.4} & \textbf{16.1} \\
            Dynamic Ckpt Avg & \textbf{4.4} & 16.6 \\
        \bottomrule
    \end{tabular}
    \caption{In a production-like setting, both techniques were able to restore the baseline overall WER, but Static Checkpoint Averaging gave better results on the targeted testset.}
    \label{tab:cf_results}
\end{table}

As shown in Table~\ref{tab:cf_results}, this approach was able improve model quality on the targeted testset, but we did observe the expected forgetting of the overall distribution. From the baseline, we saw that filtering on-device training examples to those containing targeted words allowed us to improve 8\% relative on the targeted testset. However, we also saw a 4.5\% regression on the overall WER, highlighting the necessity of techniques to mitigate catastrophic forgetting.

\subsubsection{Checkpoint Averaging}
\label{sssec:ckptavg}
Both Static Checkpoint Averaging and Dynamic Checkpoint Averaging were able to bring us back to parity on the overall distribution, while still affording improvement on the targeted distribution (Table~\ref{tab:cf_results}). In particular, with Static Checkpoint Averaging, we were able to achieve the 8\% improvement on the targeted testset from pure FL, while keeping the WER on the overall distribution from the baseline.

\subsubsection{Mixture of Centralized and Federated Training}
\label{sssec:centralized}

\begin{table}
    \centering
    \small
    \begin{tabular}{c c c}
        \toprule
        \textbf{Setup} & \textbf{Overall WER} & \textbf{Targeted WER} \\
        \midrule
            Baseline & 3.5 & 16.6 \\
            FL-only & \textcolor{red}{4.4} & \textbf{16.1} \\
            FL + Centr. Mix & \textbf{3.5} & \textbf{16.1} \\
        \bottomrule
    \end{tabular}
    \caption{After refreshing server-side training data, the baseline was much improved (3.5 WER), and regression from forgetting was much greater (20\% relative). However, mixing federated and centralized training rounds restored the baseline Overall WER while keeping FL Targeted WER wins.}
    \label{tab:cent}
\end{table}

To test the technique of mixing centralized training in with the federated rounds, we first refreshed the model with more in-domain training data. This led to a much improved baseline, and a far greater regression after our targeted fine-tuning, as shown in Table~\ref{tab:cent}. This 20\% relative gap was greater than weight averaging techniques could address. However, once we mixed in centralized training simultaneously with the federated training rounds, we were able to achieve the best of both worlds: keep the wins on Targeted WER from targeted on-device training, while achieving the same best Overall WER from the server-only baseline.

\subsection{Word Improvement}
\label{ssec:wordimprove}

\subsubsection{Error Correction Percent}
\label{sssec:ecp}

To directly understand how much fine-tuning improved model recognition of each wordlist word, we computed an Error Correction Percent according to the following formula:

\begin{equation}
EC \% = \frac{Acc_{exp} - Acc_{base}} {1.0 - Acc_{base}}
\end{equation}

Each wordlist word appears in multiple testset examples; for each word, the accuracy $Acc$ is the number of examples where the word was correctly recognized, normalized by the total number of examples containing the word. The EC\% shows how many errors made by the baseline model were corrected by the fine-tuned model, as a ratio of the total number of errors for each word. For words with an EC\% of 100\%, fine-tuning corrected all errors made by the baseline.

\begin{figure}[ht]
\centering
\includegraphics[width=.8\columnwidth,bb=0 0 808 568]{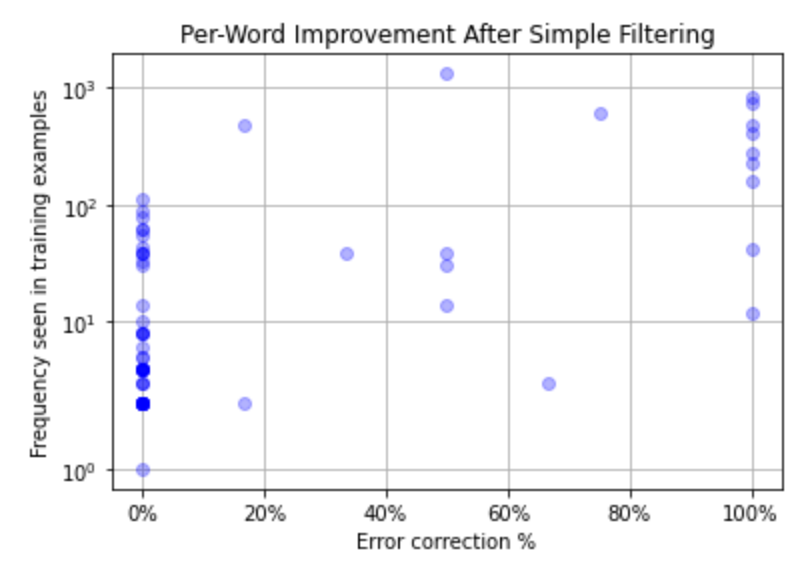}
\caption{Comparing number of errors before and after on-device fine-tuning, most words seen at least 100 times had all errors corrected after FT. This suggests that more training examples are needed for the remainder.}
\label{fig:ecp}
\end{figure}

Looking closely at how each wordlist word was improved in Figure~\ref{fig:ecp}, we saw that the majority of words that were seen at least 100 times in training had 100\% of their errors fixed by the experiment. Many words that were seen less frequently were also improved, but errors were not fixed in all their occurrences. This suggested that it was important to increase the frequency of training examples of words that we rarely saw, motivating our probabilistic sampling approach.

\subsubsection{Probabilistic Sampling and Client Loss Weighting}
\label{sssec:withprobsamp}

\begin{figure}[ht]
\centering
    \subfigure[Unique words over rounds]{
    \includegraphics[width=.45\columnwidth]{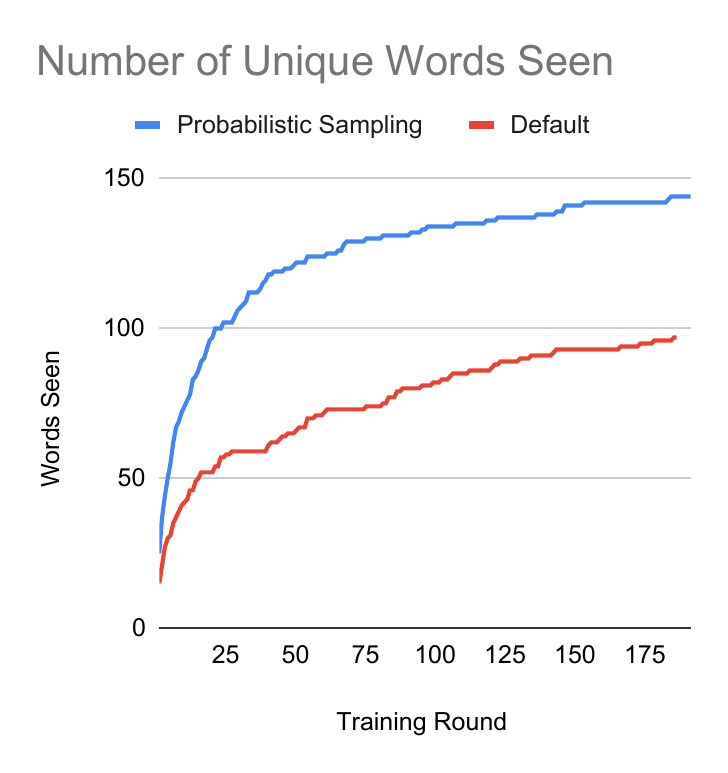}}
    \subfigure[Word count histogram]{
    \includegraphics[width=.45\columnwidth]{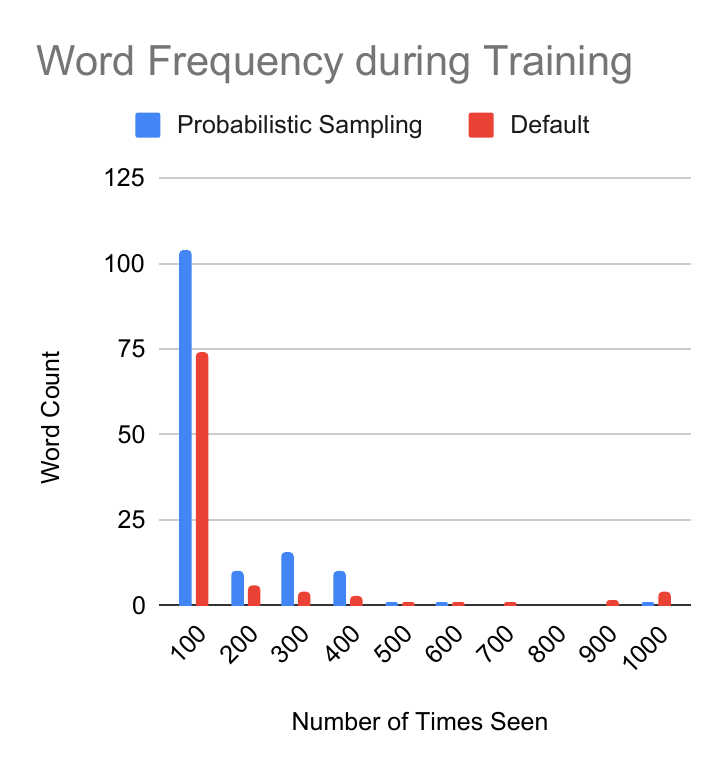}}
\caption{Probabilistic sampling was able to improve the number of unique words seen, as well as the number of words that were seen at least 100 times.}
\label{fig:prob_eet}
\end{figure}

By adding Probabilistic Sampling during fine-tuning, we increased the number of unique words seen from around 100 to closer to 150. Additionally, we increased the number of words seen at least 100 times (Fig.~\ref{fig:prob_eet}).

\begin{table}
    \centering
    \small
    \begin{tabular}{c c c}
        \toprule
        \textbf{Setup} & \textbf{Overall WER} & \textbf{Tgt WER} \\
        \midrule
            Baseline & 4.4 & 17.5 \\
            Simple Fine-Tuning & \textcolor{red}{4.6} & 16.1 \\
            Static Ckpt Avg & \textbf{4.4} & 16.1 \\
            Client Loss Weighting & \textbf{4.4} & 16.0 \\
            Probabilistic Sampling & \textcolor{red}{4.6} & \textbf{15.8} \\
            Prob Samp + St Ckpt Avg & 4.5 & 15.9 \\
        \bottomrule
    \end{tabular}
    \caption{Client Loss Weighting was able to slightly improve our already fine-tuned targeted WER from 16.1 to 16.0. Probabilistic Sampling did even better, improving our targeted WER to 15.8. The WER on the overall distribution can be restored by reintroducing checkpoint averaging.}
    \label{tab:probsamp_ckptavg}
\end{table}

As shown in Table~\ref{tab:probsamp_ckptavg}, in comparison with our previous best fine-tuned result of 16.1, using Client Loss Weighting was able to give us a small improvement to 16.0. Probabilistic Sampling gave more significant wins, bringing our targeted WER down to 15.8.

\subsubsection{Contact Names}
\label{sssec:contactnames}

As an additional application, we turned these techniques towards a related domain. Rather than learning fresh terms, we experimented with whether we could improve recognition of another class of commonly misrecognized terms: names from users' contacts lists. Names are difficult to transcribe correctly due to variety and alternate spellings, and benefit greatly from a diverse and fresh training corpus.

\begin{table}
    \centering
    \small
    \begin{tabular}{c c c}
        \toprule
        \textbf{Setup} & \textbf{Overall WER} & \textbf{Contact Names WER} \\
        \midrule
            Baseline & 3.8 & 5.7 \\
            Fresh Terms Filter & 3.8 & 5.5 \\
            Names Filter & 3.8 & \textbf{5.4} \\
        \bottomrule
    \end{tabular}
    \caption{Contact name recognition was improved by the above techniques. Using a names wordlist for filtering instead gave even better results, without harming the Overall WER.}
    \label{tab:contact_names}
\end{table}

Even using just the fresh wordlist filtering, training in FL was sufficient to improve the WER on our Contact Names testset from 5.7 to 5.5, as shown in Table~\ref{tab:contact_names}. Though we did not target any names with our filtering, the general exposure to fresh training examples was able to improve the WER. We further improved this by replacing the wordlist with a list of the top 1000 baby names in each the US, China, and India. With this new setting, we reached 5.4 WER on the testset, a 5\% relative improvement.

This demonstrates that the proposed techniques can be applied to other settings, such as learning words of a certain domain, while maintaining quality on the overall distribution.
\section{Conclusion}
\label{sec:conclusion}

In this work, we explored how user corrections can be leveraged to improve ASR model quality, particularly on fresh terms not available during model training, or on terms that the ASR model tends to get wrong. The naive approach of training directly on novel on-device user data sources posed a number of difficulties, including the challenge of targeting these inherently long-tail words, as well as catastrophic forgetting leading to degradation of model quality on the overall distribution.

To address these issues, we applied a number of techniques, such as checkpoint averaging, mixing centralized and decentralized training, and probabilistic sampling. Finally, we experimentally demonstrated the potential of this technique, both for the original problem space of learning fresh terms, as well as in adapting the model to other difficult domains, such as names. We hope that these findings may lead to further exploration of adapting ASR models to an ever-changing language corpus.
\section{Acknowledgements}
\label{sec:ack}

\ifblind
    (redacted for blind review)
\else
    We thank Yonghui Xiao, Andrew Hard, Sean Augenstein, Khe Chai Sim, Guru Prakash, and Tien-Ju Yang for their valuable contributions to this work.
\fi
\vfill\pagebreak\clearpage
\bibliographystyle{IEEEtran}
\bibliography{refs}
\end{document}